\documentclass[lettersize,journal]{IEEEtran}
\usepackage{amsmath,amsfonts}
\usepackage{algorithm}
\usepackage{array}
\usepackage[caption=false,font=normalsize,labelfont=sf,textfont=sf]{subfig}
\usepackage{textcomp}
\usepackage{stfloats}
\usepackage{url}
\usepackage{verbatim}
\usepackage{graphicx}
\usepackage{cite}
\usepackage{algorithm} 
\usepackage{textcomp}
\usepackage{url}
\usepackage{stfloats}
\usepackage{textcomp}
\usepackage{verbatim}
\usepackage{multirow}
\usepackage{array}
\usepackage{soul}  %
\usepackage{amsmath}
\usepackage{makecell}
\usepackage{booktabs}
\usepackage{colortbl}   %
\usepackage{algpseudocode}
\usepackage{hyperref}
\begin{document}

\title{A Multihead Continual Learning Framework for Fine-Grained Fashion Image Retrieval with Contrastive Learning and Exponential Moving Average Distillation}
\author{Ling Xiao,~\IEEEmembership{Senior Member,~IEEE} and Toshihiko Yamasaki,~\IEEEmembership{Senior Member,~IEEE} 
\thanks{L. Xiao is with Graduate School of Information Science and Technology, Hokkaido University, Sapporo,
Japan (e-mail: ling@ist.hokudai.ac.jp)}
\thanks{T. Yamasaki is with Graduate School of Information Science and Technology, The University of Tokyo, Tokyo, Japan (e-mail: yamasaki@cvm.t.u-tokyo.ac.jp)}
}


\maketitle

\begin{abstract}
Most fine-grained fashion image retrieval (FIR) methods assume a static setting, requiring full retraining when new attributes appear, which is costly and impractical for dynamic scenarios. Although pretrained models support zero-shot inference, their accuracy drops without supervision, and no prior work explores class-incremental learning (CIL) for fine-grained FIR.
We propose a multihead continual learning framework for fine-grained fashion image retrieval with contrastive learning and exponential moving average (EMA) distillation (MCL-FIR). MCL-FIR adopts a multi-head design to accommodate evolving classes across increments, reformulates triplet inputs into doublets with InfoNCE for simpler and more effective training, and employs EMA distillation for efficient knowledge transfer.
Experiments across four datasets demonstrate that, beyond its scalability, MCL-FIR achieves a strong balance between efficiency and accuracy. It significantly outperforms CIL baselines under similar training cost, and compared with static methods, it delivers comparable performance while using only about 30\% of the training cost. The source code is publicly available\footnote{\url{https://github.com/Dr-LingXiao/MCL-FIR}}.

\end{abstract}


\section{Introduction}
\label{sec:introduction}
Fashion plays a crucial role in shaping consumer behavior, which motivates the need to analyze visual information in fashion images. As a result, understanding and modeling visual compatibility~\cite{song2017neurostylist,song2018neural} and visual similarity~\cite{Liang_IEEE-TMM_16,Chen_IEEE-TMM_17, Liu_IEEE-TMM_20,Tang_TCSVT23,Gu_IEEE-TMM_18,Liu_CVPR_16,Dodds_arxiv_22,Goenka_CVPR_22,Sharma_WACV_21,Xiao_ICIP_22,Xiao_MIPR23_fashion,Jing_IEEE-TMM_21} has become critical research focuses~\cite{shukla2025can,zhang2025fgpr}. 
This paper goes beyond general similarity modeling and targets fine-grained fashion image retrieval (FIR), which focuses on capturing subtle visual differences between similar items~\cite{dong2025open,Veit_CVPR_17,Ma_AAAI_20,Dong_TIP21,Wan_ETAI_22,Yan_ICME_22,xiao_MIPR_23,Xiao_Access_24}. Fine-grained FIR can improve user experience by enabling more precise and relevant search results. It can also benefit the fashion industry by increasing sales and conversion rates. Moreover, fine-grained FIR protects fashion copyrights by identifying design similarities and preventing plagiarism. 

Fine-grained FIR is typically attribute-guided, where both the image and the attribute serve as query inputs. Most supervised methods therefore focus on designing attention mechanisms to extract discriminative attribute-aware features, but they operate in a static setting~\cite{Veit_CVPR_17, Ma_AAAI_20, Wan_ETAI_22, Yan_ICME_22, jiao2022fine, xiao_MIPR_23, Xiao_KBS_25, Xiao_TAI_25}. When a new attribute is introduced, the entire model must be retrained, which is impractical for real-world systems where user requirements continually evolve. For example, RPF~\cite{Dong_SIGIR_23} requires 121.77 hours to train on FashionAI using a single A100 GPU.
Another line of work relies on large-scale image–text pretraining followed by prompt tuning~\cite{han2023fashionsap}. However, these approaches mainly adapt a pretrained semantic space rather than learning how to efficiently acquire new attribute-specific visual cues. Since prompt tuning does not update the visual encoder, it struggles when entirely new attributes emerge.
These limitations highlight the need for a class-incremental learning framework that can efficiently integrate new attributes for real-world fine-grained FIR.

This paper proposes MCL-FIR, a multihead continual learning framework for fine-grained fashion image retrieval with contrastive learning and exponential moving average (EMA) distillation. As illustrated in Fig.~\ref{fig:diff}, unlike static methods, MCL-FIR updates only the new attribute while preserving performance on previously learned ones. We address three key challenges in continual fine-grained FIR. First, single-head models fail to adapt to evolving attribute distributions, so we introduce lightweight task-specific attention heads to enable stable, non-interfering updates. Second, triplet loss requires costly and unstable triplet construction, especially under incremental updates. We replace it with an InfoNCE loss on doublets, removing the need for triplet sampling while retaining strong contrastive signals. Third, continual updates to a shared encoder cause catastrophic forgetting; we mitigate this with EMA distillation that provides stable temporal supervision. 

We also implement two representative class-incremental learning (CIL) baselines for comparison. Experiments show that MCL-FIR significantly outperforms these baselines and matches state-of-the-art static methods while requiring far less computation. Our main contributions are:

\begin{figure*}[t]
\centering
\includegraphics[width=0.9\textwidth]{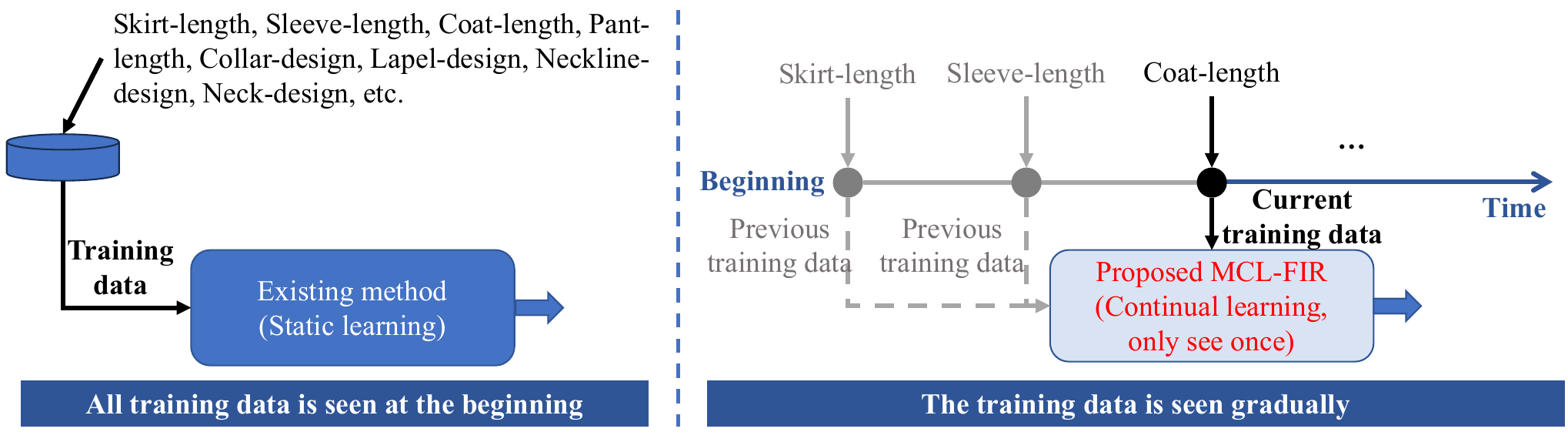}
\caption{The difference between MCL-FIR and SOTA static learning methods.}
\label{fig:diff}
\end{figure*}

\begin{figure*}[t]
\centering
\includegraphics[width=0.97\textwidth]{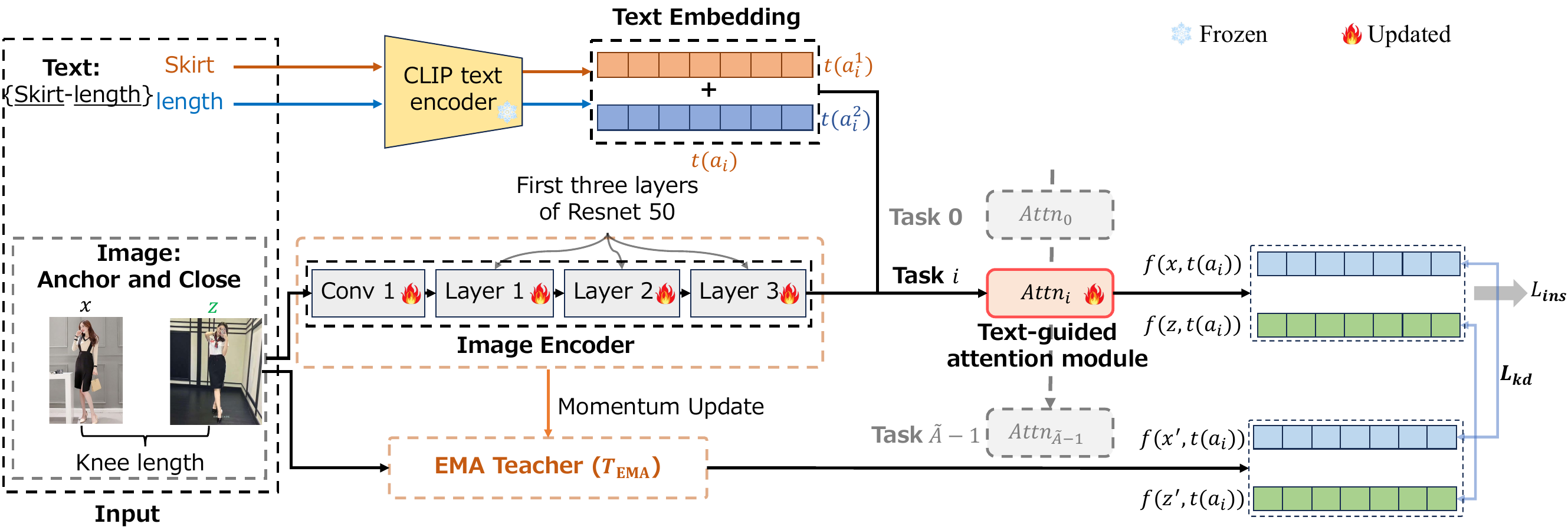}
\caption{Overall pipeline of the proposed MCL-FIR model. $Attn_{\{0\sim\tilde{A}-1\}}$ denote the respective attention module for each task (attribute). In this paper, $\widetilde{A}$ is set to $22$ to accommodate the eight attributes in FashionAI, five in DeepFashion, and nine in DARN. If the input attribute is ``skirt-length,'' it can be split into two words, with $a_i^{1}$ denoting ``skirt,'' and $a_i^{2}$ denoting ``length.'' $t(a_{i}^{1})$ and $t(a_{i}^{2})$ represent the corresponding text embeddings, extracted using a pre-trained text encoder.}
\label{fig:model}
\end{figure*}

\begin{figure*}[t]
\centering
\includegraphics[width=0.85\textwidth]{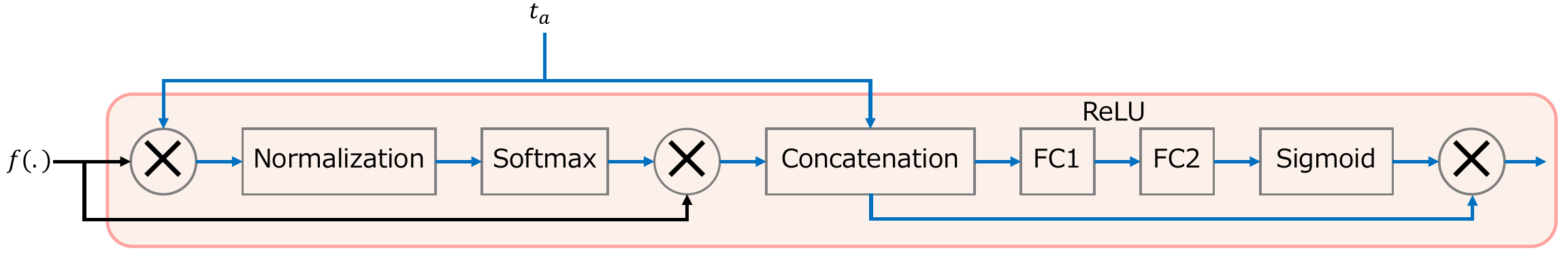}
\caption{The detailed structure of $Attn_{\{0\sim\tilde{A}-1\}}$. Note that $Attn_{0}$, $Attn_{1}$, ..., and $Attn_{\tilde{A}-1}$ share the same structure, as shown in this figure.}
\label{fig:T-attn}
\end{figure*}

\begin{enumerate}
    \item We propose a multihead CIL framework for fine-grained FIR that can incorporate new attributes without degrading performance on previously learned ones.
    \item We reduce triplet sampling to doublet sampling through the InfoNCE loss, cutting one-third of the computation, and introduce EMA-based distillation to support effective and efficient incremental updates.
    \item Compared with two implemented CIL baselines, our method is significantly more effective. Relative to state-of-the-art static FIR methods, MCL-FIR achieves a superior balance between accuracy and efficiency.
\end{enumerate}

\section{Related Works}
\subsection{Fine-grained FIR}
Fine-grained FIR faces challenges such as intra-attribute, subtle, and viewpoint variations. To address these, two main approaches have emerged: attention models for static settings and pre-trained models~\cite{huang2025fashionfae}.  
Veit {\it et al.}~\cite{Veit_CVPR_17} introduced a unified embedding space with fixed masks for selecting attribute-specific dimensions. Ma {\it et al.}~\cite{Ma_AAAI_20} extended this to multiple embedding spaces. Wan {\it et al.}~\cite{Wan_ETAI_22} fused attribute-aware channel and spatial attention, while Yan {\it et al.}~\cite{Yan_ICME_22} used iterative learning for more precise attribute localization. Jiao {\it et al.}~\cite{jiao2022fine} incorporated instance- and cluster-level supervision with online clustering. Xiao {\it et al.}~\cite{Xiao_TAI_25} applied contrastive learning to handle viewpoint variations, and Xiao {\it et al.}~\cite{Xiao_KBS_25} proposed multi-level knowledge distillation to enhance SOTA methods.
Although effective, these methods are not scalable and require retraining when attributes change, and they involve long training times, limiting their practical use. Pre-trained models such as FashionSAP~\cite{han2023fashionsap} avoid retraining, but they are not designed to learn new attributes, as prompt tuning merely adapts the pretrained semantic space without building new attribute-specific discriminative features. Therefore, a formulation that explicitly supports continual updates is needed.

\subsection{Class-Incremental Learning}
Class-Incremental Learning methods learn from a stream of data drawn from a non-stationary distribution~\cite{zhou2024class,cao2025class,yang2021learning,zhang2023c2mr}. These methods are expected to scale to a large number of tasks without incurring excessive computational or memory overhead. Their goal is to leverage knowledge from previously learned classes to facilitate the learning of new ones. During each training session, the learner has access only to data from a single task.

One of the most representative approaches is Experience Replay (ER)~\cite{rolnick2019experience}, in which a subset of past samples is replayed during the training of new tasks to mitigate catastrophic forgetting. Another common strategy is the multi-head architecture~\cite{kim2022multi}, which assigns a separate output head for each task to isolate task-specific knowledge. Other methods leverage knowledge distillation~\cite{wang2024improving,michel2024rethinking}, pretrained models~\cite{zhou2024expandable}, and regularization-based methods~\cite{huang2024class} to alleviate catastrophic forgetting and enhance model performance. Zhang {\it et al.}~\cite{zhang2023c2mr} proposed an online continual learning setup, OC-CMR, to formalize the data-incremental
growth challenge faced by cross-modal retrieval systems. However, existing CIL methods are mainly designed for classification and detection tasks, and cannot be directly applied to fine-grained fashion image retrieval (FIR), which requires learning subtle visual distinctions and preserving feature embedding consistency.
Our study bridges this gap by tailoring CIL to fine-grained FIR.

\section{Methods}
\label{sec:method}


This study addresses the limitations of static attention networks that require full retraining for new attributes, as well as pretrained models that cannot effectively handle unseen attributes. We propose MCL-FIR, a multihead continual learning framework for fine-grained fashion image retrieval with contrastive learning and EMA distillation. MCL-FIR offers the following advantages: 1) Scalability: The multi-head structure enables new attributes to be integrated without modifying previously learned components. 2) Efficiency: By reformulating triplet inputs into doublets using the InfoNCE loss, MCL-FIR simplifies training and reduces computation; in addition, lightweight attention modules support efficient incremental learning. 3) Accuracy: EMA distillation stabilizes feature representations across tasks and enhances retrieval performance.

\subsection{Overall pipeline of MCL-FIR.}
Fig.~\ref{fig:model} shows the architecture of our proposed MCL-FIR. The detailed training pipeline of MCL-FIR is described in Algorithm~\ref{alg:training_MCL-FIR}. Formally, we define a set of $\widetilde{A}$ tasks, which are learned sequentially. Each task contains a unique attribute and does not overlap with any other task. The incremental learning problem is defined as:
\begin{equation}
\mathcal{T} = \left[ (a_{0}, D_0), (a_1, D_1), \dots, (a_{\widetilde{A}-1}, D_{\widetilde{A}-1}) \right],
\end{equation}
where $a_i$ denotes the attribute associated with task $i$, $D_{i} = \{(x_0, z_0), \dots, (x_{N-1}, z_{N-1})\}$ is the corresponding training data, $N$ denotes the number of training doublets in each task, and $\widetilde{A}$ denotes the total number of attributes among all considered datasets. During training on task $i$, the learner has access only to $D_{i}$.

\begin{algorithm}[t] 
\caption{Training Procedure of MCL-FIR}
\begin{algorithmic}[1]
\State Randomly sample doublet inputs $\{x_{j}, z_{j} | a_i\}$ from $D_i$, $j \in \{0, 1, \ldots, N{-}1\}$, and $i \in \{0, 1, \ldots, \widetilde{A}{-}1\}$.
\State Initialize MCL-FIR model $\Theta$ with a shared image encoder $S$ and a text-guided attention module $Attn_i$.
\State Obtain an EMA teacher model from the image encoder, denoted as $T_{\text{EMA}}$.
\If{in training stage}
    \For{$j = 0$ to $N{-}1$}
        \State Obtain the distorted $(x^{\prime}_{j}, z^{\prime}_{j})$ using a random perspective distortion as in Ref.~\cite{Xiao_TAI_25}.
        \State Obtain the attribute embedding $t(a_i)$ using a pre-trained text encoder.
        \State Extract image features $(f(x_j), f(z_j))$ with $S$ and $(f(x^{\prime}_j), f(z^{\prime}_j))$ with the EMA teacher $T_{\text{EMA}}$.
        \State Compute attribute-aware features using $Attn_i$:
      $\{f(x_j, t(a_i)), f(z_j, t(a_i)), f(x'_j, t(a_i)), f(z'_j, t(a_i))\}$.
        \State Compute $L_{\text{ins}}$ and $L_{\text{kd}}$.
    \EndFor
    \State Update model parameters by minimizing Eq.~\ref{eq:final-loss}.
\EndIf
\end{algorithmic}
\label{alg:training_MCL-FIR}
\end{algorithm}

Specifically, we begin by randomly sampling doublet inputs $\{x_{j},z_{j}|a_i\}$ from all training datasets, where $j \in \{0,1,\ldots,N-1\}$, $i \in \{0,1,\ldots,\widetilde{A}-1\}$. The two items in a doublet input share the same attribute $a_i$ and belong to the same subclass $a_i^{\rm sub}$, 
where $a_i^{\rm sub}$ denotes a specific value of the attribute $a_i$. We applied a random perspective transformation~\cite{Xiao_TAI_25} to $(x_{j}, z_{j})$, and obtained $(x^{\prime}_{j}, z^{\prime}_{j})$.

Following this, we decompose the attribute $a_i$ into its constituent words (e.g., $a_i^{1}$ and $a_i^{2}$ for a two-word attribute) and employ a frozen pre-trained text encoder to extract the attribute features $t(a_i^{1})$ and $t(a_i^{2})$. These features are then combined using an addition operation to derive $t(a_i)$. ResNet50 is selected as the image encoder $S$. Specifically, to retain sufficient spatial features for subsequent processing, we use the output of block three in ResNet50. We also obtain an EMA teacher from the image encoder $S$, denoted as $T_{\text{EMA}}$. $S$ extracts the image features of $(x_{j}, z_{j})$, and $(f(x_{j}), f(z_{j}))$ is obtained. $T_{\text{EMA}}$ extracts the image features of $(x^{\prime}_{j}, z^{\prime}_{j})$, and $(f(x^{\prime}_{j}), f(z^{\prime}_{j}))$ is obtained.

Next, the obtained image features and the attribute feature $t(a_i)$ are processed by our attention module $Attn_i$ to generate attribute-focused image representations $\{f(x_{j}, t(a_i)), f(z_{j}, t(a_i)), f(x^{\prime}_{j}, t(a_i)), f(z^{\prime}_{j}, t(a_i))\}$. The details of $Attn_i$ will be given in Subsection~\ref{subsec:attn}. We propose replacing the traditional triplet ranking loss, which relies on triplet inputs (anchor, positive, negative), with the InfoNCE loss~\cite{rusak2024infonce} using positive pairs. The details will be provided in Subsection~\ref{subsec:InfoNCE}. Specifically, we calculate instance contrastive losses ($L_{\rm ins}$ of the pairs $(f(x_{j}, t(a_i)), f(z_{j},t(a_i)))$ and $(f(x^{\prime}_{j}, t(a_i)), f(z^{\prime}_{j},t(a_i)))$ .

We also use the MSE loss to compute the distillation loss for the embedding differences of  $(f(x_{j}), f(x^{\prime}_{j}))$ and $(f(z_{j}), f(z^{\prime}_{j}))$. These losses are added and denoted as $L_{\rm kd}$. 

Finally, the overall loss is a weighted sum of $L_{\rm ins}$ and $L_{\rm kd}$, as follows:
 \begin{equation}
 L = L_{\rm ins}  + \lambda L_{\rm kd}, 
  \label{eq:final-loss}
\end{equation}
where $\lambda$ is set as $0.0001$.

\subsection{Text-guided attention module.} 
\label{subsec:attn}
The attention module is motivated by three factors: (1) pose/scale diversity makes fixed regions impractical; 
(2) focusing on the most relevant area per attribute improves performance; 
(3) channel attention helps handle regions linked to multiple attributes by reweighting feature channels. The deatiled structure is provided in Fig.~\ref{fig:T-attn}

\noindent\textbf{1) Attribute-aware spatial attention.}
Given an input image $x_j$, the backbone extracts a feature map 
$f(x_j)\in\mathbb{R}^{1024\times h\times w}$.
We first reduce its channel dimension using a $1{\times}1$ convolution
followed by batch normalization and $\tanh$ activation:
\begin{equation}
    \tilde f(x_j)
    = \tanh\!\big(\mathrm{BN}(\mathrm{Conv}^{128}_{1\times1}(f(x_j)))\big)
    \in \mathbb{R}^{128\times h\times w}.
\end{equation}

For attribute conditioning, we take two CLIP text embeddings 
$a_i^{1},a_i^{2}\in\mathbb{R}^{512}$ for two-word attribute, sum them, and project them into 
a 128-dimensional attribute vector:
\begin{equation}
    t(a_i) = W_t\,(a_i^{1} + a_i^{2}) \in \mathbb{R}^{128}.
\end{equation}
This vector is broadcast to match the spatial size:
\begin{equation}
    T(a_i)\in\mathbb{R}^{128\times h\times w},
    \qquad
    T(a_i)_{:,h,w} = t(a_i).
\end{equation}

We compute spatial compatibility by element-wise multiplication between 
$\tilde f(x_j)$ and $T(a_i)$, summing over channels and normalizing by $1/\sqrt{128}$:
\begin{equation}
    S_{h,w}
    = \frac{1}{\sqrt{128}}
      \sum_{c=1}^{128}
      \tilde f(x_j)_{c,h,w}\,T(a_i)_{c,h,w}.
\end{equation}
A spatial attention map is obtained by applying softmax over all $(h,w)$:
\begin{equation}
    A_{1,h,w}
    = \frac{\exp(S_{h,w})}{
        \sum_{h',w'} \exp(S_{h',w'}) }
    \in \mathbb{R}^{1\times h\times w}.
\end{equation}

The attended feature map is then:
\begin{equation}
    f_{\mathrm{s}}(x_j,a_i)
    = \tilde f(x_j)\,\odot\,A
    \in \mathbb{R}^{128\times h\times w},
\end{equation}
and is aggregated across spatial locations:
\begin{equation}
    \bar f_{\mathrm{s}}
    = \sum_{h,w} f_{\mathrm{s}}(x_j,a_i)_{:,h,w}
    \in \mathbb{R}^{128}.
\end{equation}

\noindent\textbf{2) Attribute-aware channel attention.}
We reuse the attribute vector $t(a_i)\in\mathbb{R}^{128}$ and concatenate it with 
the aggregated image feature $\bar f_{\mathrm{s}}$:
\begin{equation}
    u = [\,\bar f_{\mathrm{s}};\; t(a_i)\,] 
    \in \mathbb{R}^{256}.
\end{equation}
Two fully connected layers with ReLU and sigmoid produce a channel-wise gate:
\begin{equation}
    W_{\mathrm{CA}}
    = \sigma\!\big(W_{2}\,r(W_{1}u)\big)
    \in \mathbb{R}^{128}.
\end{equation}
Finally, the attribute-modulated representation is:
\begin{equation}
    f(x_j,a_i)
    = \bar f_{\mathrm{s}} \,\odot\, W_{\mathrm{CA}}
    \in \mathbb{R}^{128}.
\end{equation}

\subsection{InfoNCE loss.}
\label{subsec:InfoNCE}
While existing static methods rely on triplet input, we adopt doublets with InfoNCE loss to enable instance-level contrastive learning. 
For each mini-batch of size $B$, we sample $B$ anchors $x_j$ and their corresponding positives $z_j$, and compute attribute-aware embeddings 
$\big(f(x_j,t(a_i)),\, f(z_j,t(a_i))\big)$. 
We then concatenate these along the batch dimension and apply $\ell_2$ normalization, yielding an embedding matrix $Z \in \mathbb{R}^{2B \times d}$. 
A similarity matrix $S$ is computed as $S = ZZ^\top$. 
Positive pairs are $(ii,\,ii+B)$ for $ii=1,\dots,B$ and $(ii,\,ii-B)$ for $ii=B+1,\dots,2B$; 
all other $2B-2$ entries in each row are treated as negatives.

\begin{equation}\label{eq:ins}
\begin{aligned}
L_{\rm ins} = -\frac{1}{2B}\Bigg(
&\sum_{ii=1}^{B} 
    \log \frac{\exp(S_{ii,\,ii+B}/\tau)}
              {\sum_{jj=1,\,jj\neq ii}^{2B} \exp(S_{ii,\,jj}/\tau)} \\
+ &\sum_{ii=B+1}^{2B} 
    \log \frac{\exp(S_{ii,\,ii-B}/\tau)}
              {\sum_{jj=1,\,jj\neq ii}^{2B} \exp(S_{ii,\,jj}/\tau)}
\Bigg),
\end{aligned}
\end{equation}
where $S_{ii,jj}$ denotes the similarity between the $ii$-th and $jj$-th normalized embeddings, 
and $\tau$ is the temperature parameter.

\subsection{EMA distillation.}
We extract an EMA teacher from the shared image encoder $S$, denoted as $T_{\text{EMA}}$, and compute the distillation loss between the outputs of $S$ and $T_{\text{EMA}}$. Since the EMA teacher is an average of the parameters of the image encoder $S$, to slightly increase the capacity gap, we apply a random perspective transformation to the original input as in~\cite{Xiao_TAI_25}. The distorted input is then processed by $T_{\text{EMA}}$. At each training step $t$, the teacher parameters are updated as:

\begin{equation}
\theta_{T_{\text{EMA}}}^{(t)} = \beta \cdot \theta_{T_{\text{EMA}}}^{(t-1)} + (1 - \beta) \cdot \theta_{S}^{(t)},
\end{equation}
where $\theta_{S}^{(t)}$ denotes the parameters of the image encoder at step $t$, and $\beta \in [0, 1)$ is a momentum coefficient. A higher $\beta$ results in smoother updates and more stable teacher behavior. In the experiments, we set $\beta$ to 0.999.
\begin{figure}[t]
\centering
\includegraphics[width=0.5\textwidth]{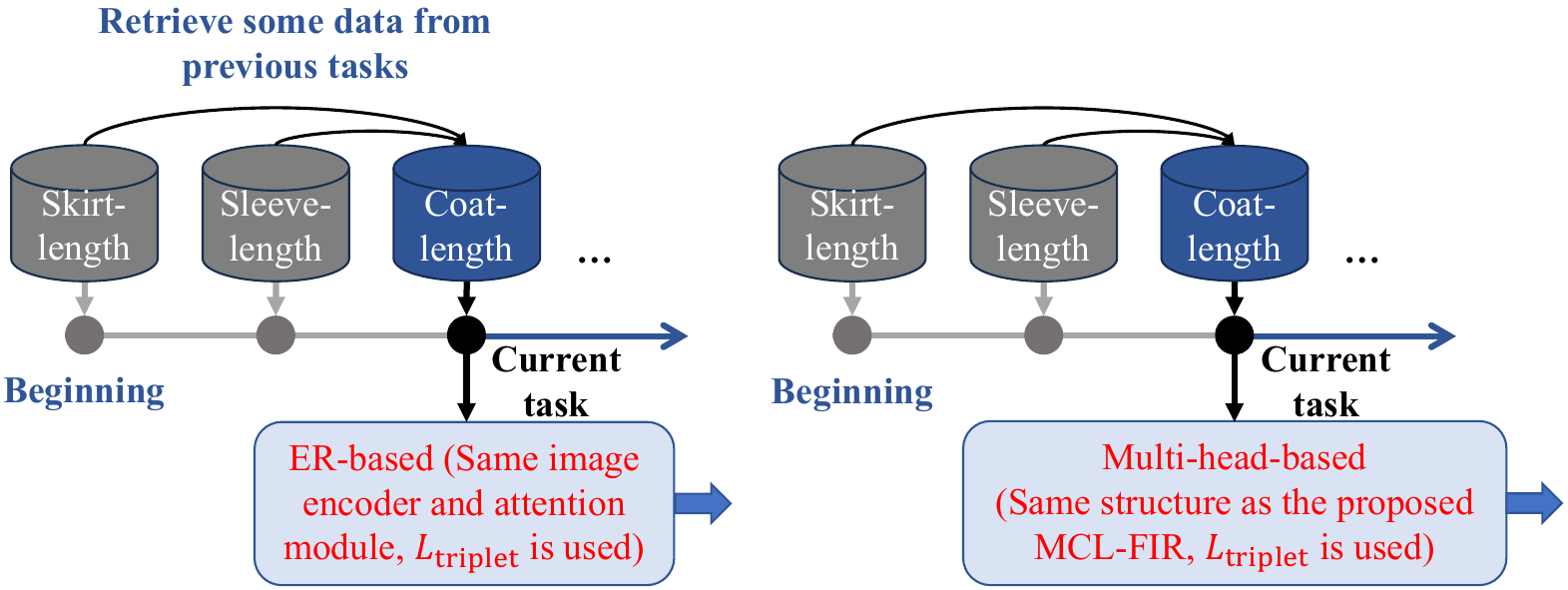}
\caption{We implemented two CIL baselines for comparison: an ER method and a multi-head method.}
\label{fig:ER-multihead}
\end{figure}

\begin{table*}[htbp]
\centering
\caption{Comparisons with the ER and multi-head CIL baselines and SOTA static methods. Results are reported as $A,(B)$, where $A$ is the final accuracy after all sequential updates and $B$ is the accuracy immediately after learning each attribute.}
\label{tab:Main_results}
\resizebox{1.0\textwidth}{!}{
\begin{tabular}{@{}lcccccccccc@{}}
\toprule
\multirow{3}{*}{} &\multirow{3}{*}{Methods} & \multicolumn{8}{c}{MAP for each attribute (Fashion AI) $\uparrow$}& \multirow{3}{*}{MAP $\uparrow$}\\ 
 & & \makecell{Skirt\\-length} & \makecell{Sleeve\\-length} & \makecell{Coat\\-length} & \makecell{Pant\\-length} &\makecell{Collar\\-design} &\makecell{Lapel\\-design} &\makecell{Neckline\\-design} &\makecell{Neck\\-design} & \\ \midrule
 \multirow{8}{*}{\makecell{Static methods}}  & CSN~\cite{Veit_CVPR_17} & 59.59 & 43.99& 42.82& 62.80&67.57 &48.10 & 41.73& 55.29& 50.80 \\  
& ASENet\_V2~\cite{Ma_AAAI_20}&64.57 &54.96 &51.76 & 64.50&71.93& 66.72& 60.29&60.83&60.76    \\ 
 & ASEN${g}$~\cite{Dong_TIP21} &   59.95& 50.72 &47.58 & 64.32 & 68.66 &59.39 &55.06  & 54.73&56.47 \\ 
& ASEN~\cite{Dong_TIP21} &  63.04  & 53.05 &52.29  & 65.26 & 71.81&61.39   &63.87  & 65.25 &61.00 \\
& ASENet\_V2+PT~\cite{xiao_MIPR_23}  &  67.50  & 60.52 & 55.20  &  70.58  &  77.35 &  72.31 & 68.31 & 67.28 & 66.29 \\ 
 & RPF~\cite{Dong_SIGIR_23}& 66.93&    69.15 & 58.83   &  72.19 &  77.14 & 72.63    &  71.48    & 71.51  &   69.38 \\  
  & ASENet\_V2+GeoDCL~\cite{Xiao_TAI_25} &  68.71  &   59.18    &   55.54    &70.72   &  77.14  &   73.03   & 68.49      & 69.25   & 66.48    \\  
  & ASENet\_V2+MKD~\cite{Xiao_KBS_25} &  69.81 &   64.22    & 61.31    & 73.86 & 78.51      &    74.10 &  70.67 & 68.70  &69.41   \\  
 
 \midrule
\multirow{3}{*}{\makecell{CIL methods}} & ER-based  & 28.66(\textcolor{blue}{43.54})   &   18.23(\textcolor{blue}{30.19})  & 21.74(\textcolor{blue}{31.69})   &    27.59(\textcolor{blue}{45.82})    &25.79(\textcolor{blue}{46.19})     & 23.62(\textcolor{blue}{40.48})   &  14.92(\textcolor{blue}{28.28})   & 25.09(\textcolor{blue}{37.49}) & 22.08(\textcolor{blue}{36.32}) \\ 
& Multi-head-based &    30.13(\textcolor{blue}{65.32})  & 17.86(\textcolor{blue}{56.36})  &  22.04(\textcolor{blue}{56.01})   &  31.99(\textcolor{blue}{67.70})   & 31.03(\textcolor{blue}{71.63})     &   27.45(\textcolor{blue}{65.94}) &   17.32(\textcolor{blue}{66.50})  &  26.05(\textcolor{blue}{60.62}) &  24.09(\textcolor{blue}{63.20})  \\
& MCL-FIR (Ours) &  64.08(\textcolor{blue}{64.08}) &   61.28(\textcolor{blue}{61.30})  &  53.25(\textcolor{blue}{53.25}) & 68.08(\textcolor{blue}{68.08}) &71.68(\textcolor{blue}{71.70})    &   70.66(\textcolor{blue}{70.67}) &   68.43(\textcolor{blue}{68.44})&  62.17(\textcolor{blue}{62.20})   &   64.41(\textcolor{blue}{64.45})  \\  
\bottomrule  
\end{tabular}}

\resizebox{0.75\textwidth}{!}{
\begin{tabular}{@{}lccccccc@{}}
\toprule
\multirow{3}{*}{}  & \multirow{3}{*}{Methods}  & \multicolumn{5}{c}{MAP for each attribute (DeepFashion) $\uparrow$}  & \multirow{3}{*}{MAP $\uparrow$}\\
&   &\makecell{Texture\\-related} & \makecell{Fabric\\-related} & \makecell{Shape\\-related} & \makecell{Pant\\-related}
 &\makecell{Style\\-related} & \\
\midrule
  \multirow{8}{*}{\makecell{Static methods}} &  CSN~\cite{Veit_CVPR_17} & 14.45 & 6.50& 11.26& 4.78& 3.47&8.07  \\
   &ASENet\_V2~\cite{Ma_AAAI_20} & 15.52  &7.19 &11.51 &5.52 &3.65 &  8.67  \\ 
  & ASEN${g}$~\cite{Dong_TIP21}  & 15.25   & 7.30 &12.09 & 5.63 & 3.84 &  8.82 \\
   & ASEN~\cite{Dong_TIP21}  & 15.01   & 7.33 & 3.48 & 6.00 &  3.77& 9.12  \\
 & ASENet\_V2+PT~\cite{xiao_MIPR_23}& 15.20   &   6.99  &   11.99 & 5.21    & 3.69 & 8.60  \\ 
  &RPF~\cite{Dong_SIGIR_23} &  16.60 &  8.82    & 14.78    &7.15     &4.95  & 10.47   \\ 
   & ASENet\_V2+GeoDCL~\cite{Xiao_TAI_25} & 15.29   &   7.11   &   11.77 &    5.52    & 3.76   & 8.68  \\  
     & ASENet\_V2+MKD~\cite{Xiao_KBS_25} & 15.94  &   7.97  &  14.10 &    6.51 &  4.14 & 9.91  \\ 
  \midrule
\multirow{3}{*}{\makecell{CIL methods}} & ER-based  &    9.42(\textcolor{blue}{12.83})    &   3.41(\textcolor{blue}{5.40})    &  5.23(\textcolor{blue}{7.81})  & 3.38(\textcolor{blue}{4.06})   &   2.43(\textcolor{blue}{2.98}) &    4.70(\textcolor{blue}{6.58})     \\ 
& Multi-head-based &    11.45(\textcolor{blue}{14.11})    &  4.47(\textcolor{blue}{7.09})     &5.73(\textcolor{blue}{11.02})    & 3.51(\textcolor{blue}{5.33})    & 2.42(\textcolor{blue}{3.53})   &    5.48(\textcolor{blue}{8.24})    \\ 
& MCL-FIR (Ours) &   15.56(\textcolor{blue}{15.57})  &  8.06(\textcolor{blue}{8.06}) &  13.38(\textcolor{blue}{13.42})   &6.61(\textcolor{blue}{6.61})    &  4.29(\textcolor{blue}{4.29})  &  9.59(\textcolor{blue}{9.61})    \\ 
\bottomrule
\end{tabular}}

\resizebox{1.0\textwidth}{!}{
\begin{tabular}{@{}lccccccccccc@{}}
\toprule
\multirow{3}{*}{}  & \multirow{3}{*}{Methods}  & \multicolumn{9}{c}{MAP for each attribute (DARN) $\uparrow$}  & \multirow{3}{*}{MAP $\uparrow$}\\
 & &\makecell{Clothes\\-category} & \makecell{Clothes\\-button} & \makecell{Clothes\\-color} & \makecell{Clothes\\-length}& \makecell{Clothes\\-pattern}& \makecell{Clothes\\-shape}& \makecell{Collar\\-shape}& \makecell{Sleeve\\-length}& \makecell{Sleeve\\-shape} & \\
\midrule
   \multirow{8}{*}{\makecell{Static methods}} & CSN~\cite{Veit_CVPR_17} & 8.16 & 23.94& 13.87 &34.89 &44.90  & 40.90 &15.20 & 66.47 & 52.37& 33.18    \\
   &ASENet\_V2~\cite{Ma_AAAI_20}& 7.75 &24.42 & 15.52 &34.38 & 44.99 & 40.56 & 15.11& 67.05 & 52.74&   33.38 \\ 
   &ASEN${g}$~\cite{Dong_TIP21}  & 7.44 & 23.68& 11.34 & 32.94& 45.00 & 40.03 &15.57 & 65.67 &54.08 &32.62    \\
    &ASEN~\cite{Dong_TIP21}  &7.48  &23.38&12.31  &31.93 &47.44  &  38.98&14.92 &66.66 & 54.56& 32.81   \\
  &ASENet\_V2+PT~\cite{xiao_MIPR_23}& 7.78 &24.73 & 15.17 & 34.14& 46.12 &40.62  & 15.22  & 68.35& 52.83 & 33.65  \\
  &RPF~\cite{Dong_SIGIR_23}&  9.91& 30.08&16.17  &41.98 & 48.90 &46.63 & 18.96& 76.16 & 56.03&38.06    \\ 
  &ASENet\_V2+GeoDCL~\cite{Xiao_TAI_25}& 7.83 &24.74 & 15.20 & 34.16& 46.10 &40.65  & 15.25  & 68.40& 52.89 & 33.73  \\
  &ASENet\_V2+MKD~\cite{Xiao_KBS_25}&  9.76& 29.15 &16.23 &40.1 & 45.7 &44.53& 18.00& 75.01 & 54.39& 36.10    \\
  \midrule
  \multirow{3}{*}{\makecell{CIL methods}} 
  &ER-based  &6.60(\textcolor{blue}{7.86})    &23.08(\textcolor{blue}{23.70})   & 11.22(\textcolor{blue}{13.31})  &   31.76(\textcolor{blue}{33.55})  &   44.26(\textcolor{blue}{42.60}) &  39.75(\textcolor{blue}{39.25})   & 14.24(\textcolor{blue}{15.00})  &  66.68(\textcolor{blue}{66.68}) & 53.06(\textcolor{blue}{52.26})  &    32.06(\textcolor{blue}{32.44})     \\  
  &Multi-head-based  &  7.42(\textcolor{blue}{10.02})  &22.34(\textcolor{blue}{25.65})   &  11.67(\textcolor{blue}{15.42})  & 31.21(\textcolor{blue}{34.55})    &  43.96(\textcolor{blue}{47.33})  &    38.59(\textcolor{blue}{39.49}) &  14.34(\textcolor{blue}{15.19})&   65.26(\textcolor{blue}{69.26}) &  51.83(\textcolor{blue}{53.23})&    31.62(\textcolor{blue}{34.18})      \\  
  &MCL-FIR (Ours)&  9.93(\textcolor{blue}{9.98}) &    29.90(\textcolor{blue}{29.93})  &   17.55(\textcolor{blue}{17.58})&  42.44(\textcolor{blue}{42.44}) &   49.97(\textcolor{blue}{49.98}) & 46.80(\textcolor{blue}{46.81})      &  18.76(\textcolor{blue}{18.77})&  75.10(\textcolor{blue}{75.10}) &    55.83(\textcolor{blue}{55.83}) & 38.23(\textcolor{blue}{38.25}) \\    
\bottomrule
\end{tabular}}

\vspace{0.2cm}
\centering
\caption{The training time when achieving performance presented in Table~\ref{tab:Main_results}.}
\label{tab:time}
\centering
\resizebox{0.6\textwidth}{!}{
\begin{tabular}{@{}lccccc@{}}
\toprule
\multirow{3}{*}{} &\multirow{2}{*}{Methods} & \multicolumn{3}{c}{Training time (h) $\downarrow$} &\multirow{2}{*}{Total training time (h) $\downarrow$} \\
& & FashionAI & DeepFashion & DARN  & \\
\midrule
\multirow{8}{*}{\makecell{Static methods}} &CSN~\cite{Veit_CVPR_17}& 12.50 & 21.00 &41.73  & 75.23  \\ 
 &ASENet\_V2~\cite{Ma_AAAI_20}& 14.00  & 21.80 & 41.27 & 77.07\\
 &ASEN${g}$~\cite{Dong_TIP21}& 30.50  & 49.72 & 74.95
 & 155.17   \\
 &ASEN~\cite{Dong_TIP21}&  47.60   & 75.22 & 203.18 & 326.00 \\
& ASENet\_V2+PT~\cite{xiao_MIPR_23} &   52.50  &  74.17 &75.75 & 202.42   \\
 &RPF~\cite{Dong_SIGIR_23}& 121.77 & 179.63 & 168.46
 & 469.86  \\
 &ASENet\_V2+GeoDCL~\cite{Xiao_TAI_25} &   52.50  &  74.16 &75.75 & 202.41   \\
 &ASENet\_V2+MKD~\cite{Xiao_KBS_25} &   53.60  &  76.10 &80.00 & 209.7  \\
  \midrule
 \multirow{3}{*}{\makecell{CIL methods}} 
&ER-based&  - &  -  &- & 84.21\\
&Multi-head-based&  - &  -  &- & \textbf{47.35}\\
&MCL-FIR (Ours)&  - &  -  &- & \underline{65.32}\\
 \bottomrule
\end{tabular}}
\end{table*}

\section{Experiments}
\label{sec:exp}

\subsection{Experimental settings}
\label{sec:exp-setting}

\noindent\textbf{Datasets.}
We evaluate fine-grained FIR on three datasets: FashionAI~\cite{Veit_CVPR_17}, DeepFashion~\cite{Liu_CVPR_16}, and DARN~\cite{Huang_ICCV_15}. Each dataset contains multiple attributes, and each attribute is treated as an individual task in the CIL setting. The three datasets are learned sequentially to mimic a challenging real-world scenario.
\textit{FashionAI}: 180,335 images with eight attributes, split 8:1:1 into 144k training, 18k validation, and 18k test images.
\textit{DeepFashion}: 289,222 images with six attributes and 1,050 sub-classes, split 8:1:1; the validation/test sets are further divided into query and candidate sets at a 1:4 ratio.
\textit{DARN}: Nine attributes, but incomplete image links create data sparsity. All methods are trained and evaluated using the same splits for fair comparison.

\begin{figure*}[htbp]
    \centering
        \includegraphics[width=1.0\textwidth]{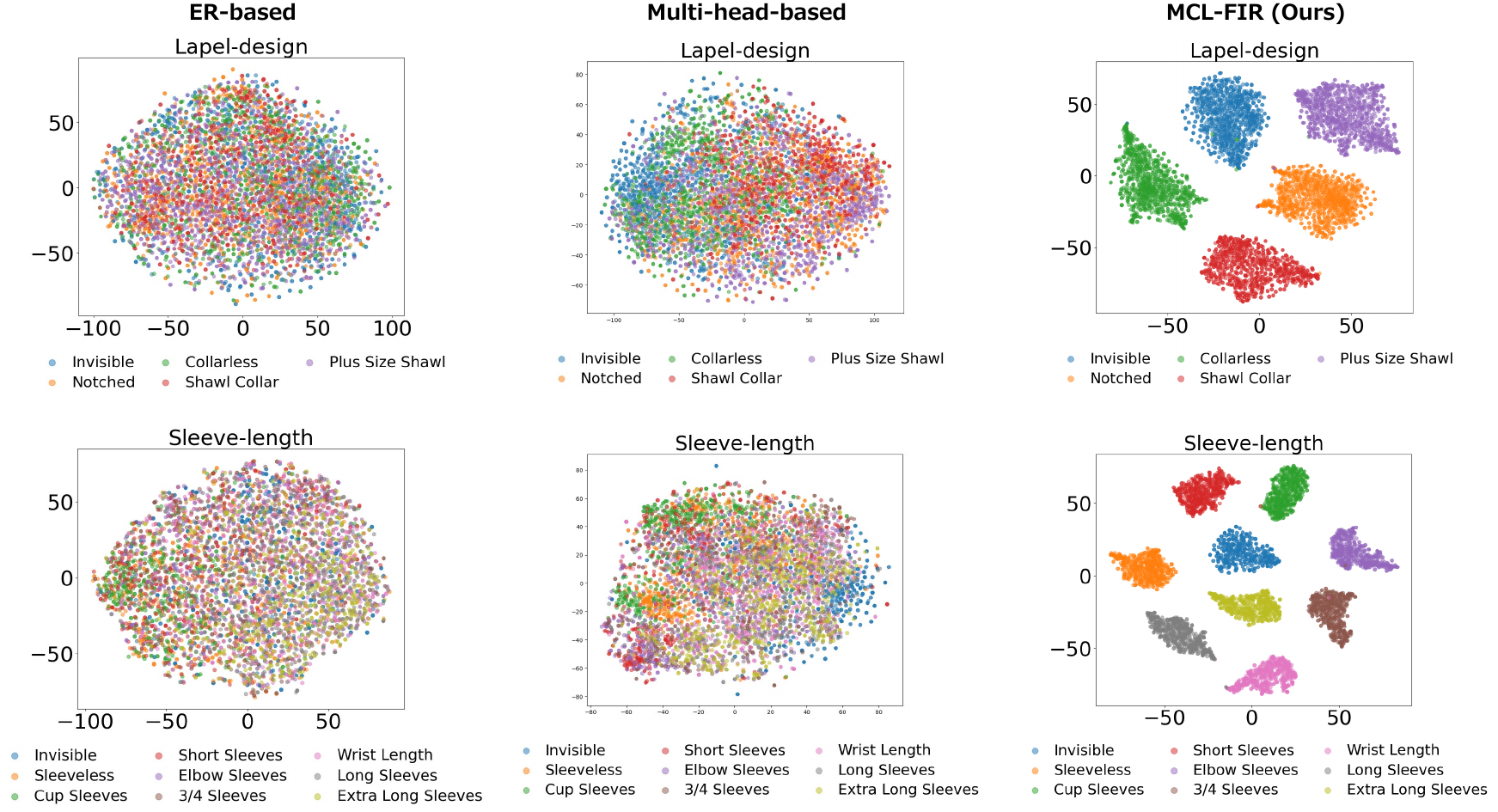}
        \caption{The t-SNE visualization on the FashionAI dataset shows that MCL-FIR effectively captures subtle differences between sub-classes within each attribute.}
        \label{fig:TSNE}
    \vspace{0.2cm}
    \centering
\includegraphics[width=1.0\textwidth]{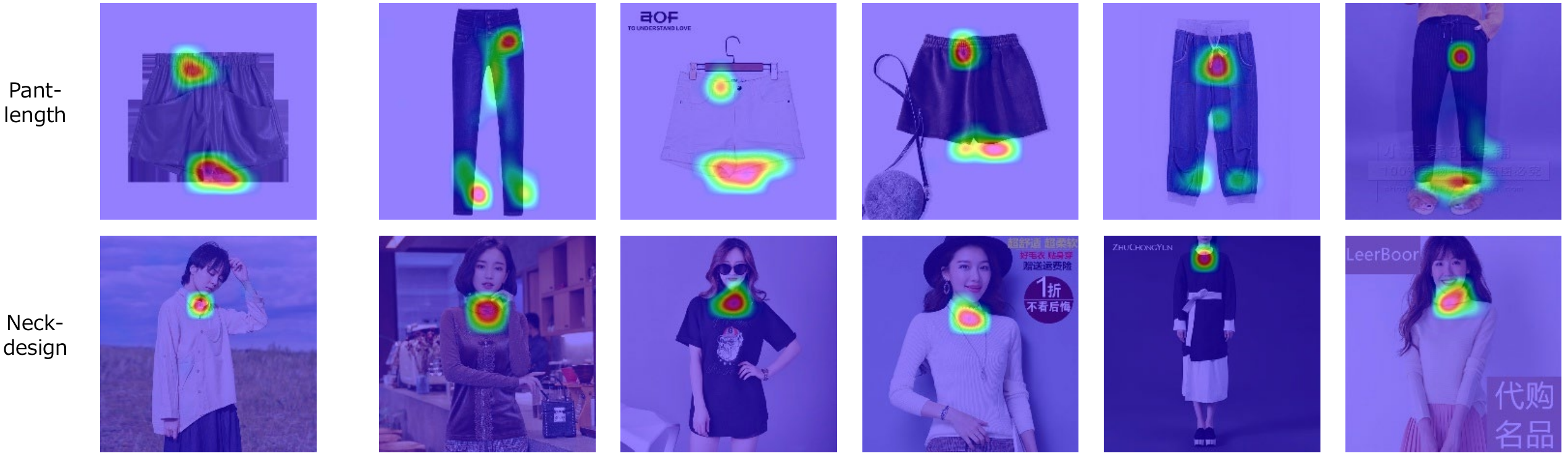}
\caption{Visualization of spatial attention guided by a specified query attribute, shown above each original image.
For length-related attributes, MCL-FIR precisely identifies the start and end regions of the pant, improving the model’s perception of length. For design-related attributes, MCL-FIR accurately highlights the corresponding regions, demonstrating strong semantic alignment.}
\label{fig:Attn}
\end{figure*}

\begin{table*}[t]
\caption{Ablation studies conducted with 10 training epochs and 20,000 training doublets.}
\begin{center}
\resizebox{0.9\textwidth}{!}{
\begin{tabular}{@{}lccccccccc@{}}
\toprule
\multirow{3}{*}{Methods}& \multicolumn{8}{c}{MAP for each attribute (Fashion AI)}& \multirow{3}{*}{MAP $\uparrow$} \\
&\makecell{Skirt\\-length} & \makecell{Sleeve\\-length} & \makecell{Coat\\-length} & \makecell{Pant\\-length} &\makecell{Collar\\-design} &\makecell{Lapel\\-design} &\makecell{Neckline\\-design} &\makecell{Neck\\-design} &  \\
\midrule
Multi-head-based ($L_{\rm triplet}$) &  29.02  &    17.28  &22.03   &30.99      &   30.93  &  27.11    &  17.31   &  26.28    &  23.72    \\   
Multi-head-based$^{+}$ ($L_{\rm triplet}$ + $L_{\text{kd}}$) &  60.80   &  51.05    & 49.56  &  64.51     &    67.69  &     65.56  &   60.99   &  60.44     & 58.92     \\   
\midrule
MCL-FIR($L_{\text{ins}}$)&  31.16  &    19.92  &   21.21 &   31.89  &    29.80   &  28.37    &  16.54  &   26.52    &  24.27     \\   
MCL-FIR($L_{\text{ins}}$ + $L_{\text{kd}}$)&   64.28     &  58.68    &  53.28   &   68.56   &  70.75    &   68.17 &     67.65  &  61.02  &  63.52   \\   
\bottomrule
\end{tabular}}
\resizebox{0.65\textwidth}{!}{
\begin{tabular}{@{}lcccccc@{}}
\toprule
\multirow{3}{*}{Methods}& \multicolumn{5}{c}{MAP for each attribute (DeepFashion)}  & \multirow{3}{*}{MAP $\uparrow$}\\
  &\makecell{Texture\\-related} & \makecell{Fabric\\-related} & \makecell{Shape\\-related} & \makecell{Pant\\-related}
 &\makecell{Style\\-related} & \\
\midrule
Multi-head-based ($L_{\rm triplet}$) &  11.66    &  4.58  &  5.88 &  3.51  &  2.44  &   5.58    \\  
Multi-head-based$^{+}$ ($L_{\rm triplet}$ + $L_{\text{kd}}$) &   14.42   &  6.90  & 10.64  &4.84   & 3.16    &  8.01    \\  
\midrule
MCL-FIR($L_{\text{ins}}$) &   12.47    & 4.88  &    6.47  &   3.61   & 3.23    & 6.07       \\  
MCL-FIR($L_{\text{ins}}$ + $L_{\text{kd}}$) &    15.68   &  7.72 &  13.70    & 6.46&  4.09   &   9.53     \\  
 \bottomrule
\end{tabular}}
\resizebox{0.9\textwidth}{!}{
\begin{tabular}{@{}lcccccccccc@{}}
\toprule
\multirow{3}{*}{Methods}   & \multicolumn{9}{c}{MAP for each attribute (DARN) $\uparrow$}  & \multirow{3}{*}{MAP $\uparrow$} \\
 &\makecell{Clothes\\-category} & \makecell{Clothes\\-button} & \makecell{Clothes\\-color} & \makecell{Clothes\\-length}& \makecell{Clothes\\-pattern}& \makecell{Clothes\\-shape}& \makecell{Collar\\-shape}& \makecell{Sleeve\\-length}& \makecell{Sleeve\\-shape} & \\
\midrule
Multi-head-based ($L_{\rm triplet}$) &   6.95  &      22.25  &  12.38  &  31.97  & 43.04    &  39.37   &  14.38 &62.92&   52.38     & 31.52    \\  
Multi-head-based$^{+}$ ($L_{\rm triplet}$ + $L_{\text{kd}}$ ) &  9.06  &     25.99  &  15.43  &  33.13  & 47.01    &  38.75   & 14.66 &69.25 &  51.74    &33.62   \\  
\midrule
MCL-FIR($L_{\text{ins}}$)  &  7.30  &  23.66     &13.34    &   33.13  &  46.75   &  40.02    & 14.87  &   67.40 &    52.62   &  33.00   \\  
MCL-FIR($L_{\text{ins}}$ + $L_{\text{kd}}$) &  10.14 &    28.62  &   15.79 &  39.42    &   48.26  & 42.21      &  18.35  &   73.84 &    53.78   &   36.39 \\  

\bottomrule
\end{tabular}}
\end{center}
\label{tab:ablation}
\vspace{0.2cm}
\centering
\caption{Experimental analysis with different dataset orders (10 training epochs and 20,000 training doublets).}
\label{tab:task-order}
\resizebox{0.9\textwidth}{!}{
\begin{tabular}{@{}lccccccccc@{}}
\toprule
\multirow{3}{*}{Methods} & \multicolumn{8}{c}{MAP for each attribute (Fashion AI) $\uparrow$}& \multirow{3}{*}{MAP $\uparrow$} \\ 
& \makecell{Skirt\\-length} & \makecell{Sleeve\\-length} & \makecell{Coat\\-length} & \makecell{Pant\\-length} &\makecell{Collar\\-design} &\makecell{Lapel\\-design} &\makecell{Neckline\\-design} &\makecell{Neck\\-design} &  \\ \midrule
FashionAI→DeepFashion→DARN &   64.28     &  58.68    &  53.28   &   68.56   &  70.75    &   68.17 &     67.65  &  61.02  &  63.52   \\  

FashionAI→DARN→DeepFashion  & 64.02   &  60.16 & 52.47  & 66.74      & 70.33 &    70.02 &  66.11  &   61.53 & 63.26    \\

DARN→DeepFashion→FashionAI &  64.92   & 60.50 &   52.58&  66.13   &   70.17 &  67.78 &  66.23 &  62.13   &  63.24   \\

DARN→FashionAI→DeepFashion &  63.17   &   58.83  &  52.49    &  65.89      & 69.32    &    66.72  &   66.80   &  60.35   & 62.52  \\

DeepFashion→DARN→FashionAI  & 64.27   &   60.29  &  52.84   &  66.07    & 70.38      &  68.94   &  67.83    &  61.15   & 63.51   \\

DeepFashion→FashionAI→DARN   & 63.13 &    60.23&  54.20   &  65.93   &  69.37   & 70.42   &   67.80   & 59.95   &  63.48   \\
 \bottomrule
\end{tabular}}
\vspace{0.2cm}
\caption{Experimental analysis of the effect of varying $\lambda$ in Eq.~\ref{eq:final-loss}, conducted over 10 epochs with 20,000 training doublets.}
\begin{center}
\resizebox{0.8\textwidth}{!}{
\begin{tabular}{@{}lcccccccccc@{}}
\toprule
\multirow{3}{*}{Methods}&\multirow{3}{*}{$\lambda$} & \multicolumn{8}{c}{MAP for each attribute (Fashion AI)}& \multirow{3}{*}{MAP $\uparrow$} \\
 &  &\makecell{Skirt\\-length} & \makecell{Sleeve\\-length} & \makecell{Coat\\-length} & \makecell{Pant\\-length} &\makecell{Collar\\-design} &\makecell{Lapel\\-design} &\makecell{Neckline\\-design} &\makecell{Neck\\-design} &  \\
\midrule
 \multirow{6}{*}{\makecell{MCL-FIR\\ (Ours)}} &0.00001  & 64.32       & 59.53     &  53.51   &    67.94  &   69.88  & 68.65   &   66.28   &  62.04   &  63.40   \\ 
 &0.0001  &   64.28     &  58.68    &  53.28   &   68.56   &  70.75    &   68.17 &     67.65  &  61.02  &  63.52   \\ 
  &0.001 &     63.35   &   59.19  &   52.97  &    67.42  &  70.43    &  67.77  &  67.89   &60.81   &   63.29 \\ 
  &0.01   &   63.42    &     59.62 &   53.29  & 66.63     & 68.47     &  69.59 &     66.56 &  62.57  &   63.16  \\ 
  & 0.1 &    62.71    &  59.52   &   54.61  &  66.41    &   69.01  &  68.65 &     67.43 &  63.00   &  63.40   \\ 
    &1 &    64.61    &  59.30   &   53.95  &   67.07   & 69.54     &  69.58  &      66.41&  58.87   &  63.19   \\ 
 \bottomrule
\end{tabular}}
\end{center}
\label{tab:lamda}
\vspace{0.2cm}
\end{table*}

\noindent\textbf{Compared methods.}
Because no existing CIL framework targets fine-grained FIR, we implement two representative CIL baselines: experience replay (ER) and a multi-head model (Fig.~\ref{fig:ER-multihead}). The multi-head baseline is trained with the triplet loss $L_{\rm triplet}$ (Eq.~\ref{eq:triplet}). We also compare MCL-FIR with SOTA static FIR methods, including CSN~\cite{Veit_CVPR_17}, ASENet\_V2~\cite{Ma_AAAI_20}, ASEN${g}$~\cite{Dong_TIP21}, ASEN~\cite{Dong_TIP21}, ASENet\_V2+PT~\cite{xiao_MIPR_23}, RPF~\cite{Dong_SIGIR_23}, ASENet\_V2+GeoDCL~\cite{Xiao_TAI_25}, and ASENet\_V2+MKD~\cite{Xiao_KBS_25}, using official code for all methods. Static approaches are trained separately on FashionAI, DeepFashion, and DARN to avoid cross-dataset conflicts.
\begin{equation}\label{eq:triplet}
    L_{\rm triplet} = \max\{0, m \!+\! S(x_j,z_j)\!-\! S(x_j,y_j)\}, \\
\end{equation}
where $\{x_j, y_j, z_j\}$ constitutes a triplet, $(x_j, y_j)$ forms a negative pair, and $(x_j, z_j)$ is a positive pair. The margin $m$ is set to 0.2. $S(x_j, z_j)$ is calculated as $ 
S(x_j,z_j) = \sum_{j} (f(x_j, t(a_i)) \cdot f(z_j,t(a_i)))$ 

Prompt-based and parameter-efficient tuning methods rely on large pretrained multimodal models and focus on single-task adaptation within a fixed representation space. Because the visual encoder remains frozen, these methods cannot acquire new attribute-specific visual cues, nor do they include mechanisms to mitigate catastrophic forgetting. Therefore, they are unsuitable for continual fine-grained FIR, and we consider them fundamentally different in scope rather than directly comparable.

\begin{table*}[htbp]
\centering
\caption{Analysis on the impact of different batch sizes during training, conducted over 10 epochs with 20,000 training doublets.}
\begin{center}
\resizebox{0.78\textwidth}{!}{
\begin{tabular}{@{}lcccccccccc@{}}
\toprule
\multirow{3}{*}{Methods}&\multirow{3}{*}{\#Batch size} & \multicolumn{8}{c}{MAP for each attribute (Fashion AI)}&\multirow{3}{*}{MAP $\uparrow$} \\
 & &\makecell{Skirt\\-length} & \makecell{Sleeve\\-length} & \makecell{Coat\\-length} & \makecell{Pant\\-length} &\makecell{Collar\\-design} &\makecell{Lapel\\-design} &\makecell{Neckline\\-design} &\makecell{Neck\\-design}  &  \\
\midrule
 \multirow{5}{*}{\makecell{MCL-FIR\\ (Ours)}}   &4 &  56.87    & 49.10    &  46.63   &  58.96   &   62.49   &   55.40   &  54.86    &   53.28  &54.01   \\ 
  &8 &    62.37  &  58.13   & 52.87    & 64.94    &  67.46  &   64.85   &   63.65   &  58.41   &  61.16 \\ 
    &16 &    63.72  & 58.29    &  52.94   & 66.82   & 69.84     & 67.56    &  67.01    &  63.26   &63.04   \\ 
    &32  &   64.28     &  58.68    &  53.28   &   68.56   &  70.75    &   68.17 &     67.65  &  61.02  &  63.52   \\  
    &64  &   61.05  &   58.92  &    51.93&    65.89 &  69.74    &   67.78   &    67.07  &    60.57 & 62.40 \\ 
 \bottomrule
\end{tabular}}
\end{center}
\label{tab:train-batch}
\vspace{0.2cm}
\caption{Performance variation with different training doublet sizes, using 10 training epochs.}
\begin{center}
\resizebox{0.78\textwidth}{!}{
\begin{tabular}{@{}lcccccccccc@{}}
\toprule
\multirow{2}{*}{Methods}&\multirow{3}{*}{\makecell{\#Training \\ doublet}} & \multicolumn{8}{c}{MAP for each attribute (Fashion AI)}&\multirow{3}{*}{MAP $\uparrow$} \\
 & &\makecell{Skirt\\-length} & \makecell{Sleeve\\-length} & \makecell{Coat\\-length} & \makecell{Pant\\-length} &\makecell{Collar\\-design} &\makecell{Lapel\\-design} &\makecell{Neckline\\-design} &\makecell{Neck\\-design} &  \\
\midrule
\multirow{6}{*}{\makecell{MCL-FIR\\ (Ours)}} &2,000& 57.36      &   46.79  &    45.55 &58.05       &    63.25  & 54.37   & 50.15   &   55.84   &    52.65  \\ 
& 5,000 &    63.32    &  52.23   & 50.98    &    63.40   &   67.44   & 63.87  &60.26   &   61.33   & 59.27     \\ 
&10,000 &  62.90     &    57.65 &    52.84 &   67.27    & 67.68     & 67.84   & 65.51    &62.92     & 62.37     \\ 
&20,000&   64.28     &  58.68    &  53.28   &   68.56   &  70.75    &   68.17 &     67.65  &  61.02  &  63.52   \\  
&50,000&     64.00   & 58.83    &  52.35   &     67.18  &    72.07  & 70.09   & 66.88   &   62.48   &   63.45   \\ 
&100,000 &    61.28    &  61.35   & 54.48    & 65.84     &    70.66  &67.48    &    65.82 &   61.07  &    63.07  \\ 
\bottomrule
\end{tabular}}
\end{center}
\label{tab:train-triplets}
\vspace{0.2cm}
\centering
\caption{Experimental results illustrating the impact of varying the number of training epochs, conducted with 20,000 training doublets.}
\label{tab:train-epochs}
\resizebox{0.8\textwidth}{!}{
\begin{tabular}{@{}lcccccccccc@{}}
\toprule
\multirow{3}{*}{Methods} &  \multirow{3}{*}{\# Epochs} & \multicolumn{8}{c}{MAP for each attribute (Fashion AI) $\uparrow$}& \multirow{3}{*}{MAP $\uparrow$} \\ 
&  & \makecell{Skirt\\-length} & \makecell{Sleeve\\-length} & \makecell{Coat\\-length} & \makecell{Pant\\-length} &\makecell{Collar\\-design} &\makecell{Lapel\\-design} &\makecell{Neckline\\-design} &\makecell{Neck\\-design} &  \\ \midrule

\multirow{5}{*}{MCL-FIR (Ours)} & 50  & 64.08&   61.28  &  53.25 & 68.08 &71.68    &   70.66 &   68.43&  62.17   &   64.41  \\
  & 40  &    63.81   &   61.34  & 53.90    &  67.81      &   71.32  &    69.37   &   68.46   &    62.11   &  64.40    \\
   & 30 &    63.40   &   62.07  & 54.41    &  67.02      &   70.78  &    69.27   &   68.49    &    62.39   &  64.35     \\
 & 20    &    64.25  &  60.40   & 53.31   &    66.46     &    69.23 &    68.55   &     67.55   &   63.26    &  63.61      \\
 & 10   &   64.28     &  58.68    &  53.28   &   68.56   &  70.75    &   68.17 &     67.65  &  61.02  &  63.52   \\

 \bottomrule
\end{tabular}}
\vspace{0.2cm}
\centering
\caption{EMA without distillation, conducted with 20,000 training doublets, 10 epochs.}
\label{tab:ema-alone}
\resizebox{0.8\textwidth}{!}{
\begin{tabular}{@{}lccccccccc@{}}
\toprule
\multirow{3}{*}{Methods} & \multicolumn{8}{c}{MAP for each attribute (Fashion AI) $\uparrow$}& \multirow{3}{*}{MAP $\uparrow$} \\ 
& \makecell{Skirt\\-length} & \makecell{Sleeve\\-length} & \makecell{Coat\\-length} & \makecell{Pant\\-length} &\makecell{Collar\\-design} &\makecell{Lapel\\-design} &\makecell{Neckline\\-design} &\makecell{Neck\\-design} &  \\ \midrule
EMA without distillation & 50.59 &  37.07  &  36.66 &  51.45 &  53.35 & 44.88   &37.87   &  43.43   & 43.07   \\
MCL-FIR (Ours) &   64.28     &  58.68    &  53.28   &   68.56   &  70.75    &   68.17 &     67.65  &  61.02  &  63.52   \\
 \bottomrule
\end{tabular}}
\vspace{0.2cm}
\centering
\caption{Continual Learning on the Zappos50K Dataset. Results are reported as $A\,(B)$, where $A$ denotes the final accuracy after sequential training and $B$ 
denotes the accuracy immediately after learning the corresponding attribute.}
\label{tab:Zappos50K_results}
\resizebox{1.0\textwidth}{!}{
\begin{tabular}{@{}lccccccccc@{}}
\toprule
\multirow{3}{*}{Methods} & \multicolumn{8}{c}{MAP for each attribute (Fashion AI) $\uparrow$}& \multirow{3}{*}{MAP $\uparrow$}\\ 
& \makecell{Skirt\\-length} & \makecell{Sleeve\\-length} & \makecell{Coat\\-length} & \makecell{Pant\\-length} &\makecell{Collar\\-design} &\makecell{Lapel\\-design} &\makecell{Neckline\\-design} &\makecell{Neck\\-design} & \\ \midrule
MCL-FIR (Ours) &  64.53(\textcolor{blue}{64.08}) &   59.52(\textcolor{blue}{61.30})  &  54.12(\textcolor{blue}{53.25}) & 66.95(\textcolor{blue}{68.08}) &70.70(\textcolor{blue}{71.70})    &  68.32(\textcolor{blue}{70.67}) &  68.62(\textcolor{blue}{68.44})&  62.71(\textcolor{blue}{62.20})   &   63.95(\textcolor{blue}{64.45})  \\  
\bottomrule  
\end{tabular}}
\resizebox{0.6\textwidth}{!}{
\begin{tabular}{@{}lcccccc@{}}
\toprule
 \multirow{3}{*}{Methods}  & \multicolumn{5}{c}{MAP for each attribute (DeepFashion) $\uparrow$}  & \multirow{3}{*}{MAP $\uparrow$}\\
 &\makecell{Texture\\-related} & \makecell{Fabric\\-related} & \makecell{Shape\\-related} & \makecell{Pant\\-related}
 &\makecell{Style\\-related} & \\
\midrule
MCL-FIR (Ours) &  15.60(\textcolor{blue}{15.57})  &  8.04(\textcolor{blue}{8.06}) & 13.19(\textcolor{blue}{13.42})   &6.59(\textcolor{blue}{6.61})    & 4.09(\textcolor{blue}{4.29})  &  9.52(\textcolor{blue}{9.61})    \\ 
\bottomrule
\end{tabular}}
\resizebox{1.0\textwidth}{!}{
\begin{tabular}{@{}lcccccccccc@{}}
\toprule
 \multirow{3}{*}{Methods}  & \multicolumn{9}{c}{MAP for each attribute (DARN) $\uparrow$}  & \multirow{3}{*}{MAP $\uparrow$}\\
&\makecell{Clothes\\-category} & \makecell{Clothes\\-button} & \makecell{Clothes\\-color} & \makecell{Clothes\\-length}& \makecell{Clothes\\-pattern}& \makecell{Clothes\\-shape}& \makecell{Collar\\-shape}& \makecell{Sleeve\\-length}& \makecell{Sleeve\\-shape} & \\
\midrule
MCL-FIR (Ours)&  10.09(\textcolor{blue}{9.98}) &   30.22(\textcolor{blue}{29.93})  &   17.53(\textcolor{blue}{17.58})&  42.74(\textcolor{blue}{42.44}) &  49.49(\textcolor{blue}{49.98}) & 46.76(\textcolor{blue}{46.81})      &  18.57(\textcolor{blue}{18.77})&  75.69(\textcolor{blue}{75.10}) &   55.86(\textcolor{blue}{55.83}) & 38.29(\textcolor{blue}{38.25}) \\    
\bottomrule
\end{tabular}}

\resizebox{0.58\textwidth}{!}{
\begin{tabular}{@{}lcccccccccc@{}}
\toprule
\multirow{3}{*}{Methods}  & \multicolumn{4}{c}{MAP for each attribute (Zappos50k) $\uparrow$}  & \multirow{3}{*}{MAP $\uparrow$}\\
 & Boots & Sandals & Shoes & Slippers & \\
\midrule
MCL-FIR (Ours)& 56.39(\textcolor{blue}{56.39}) &    82.90(\textcolor{blue}{82.90})  &  69.30(\textcolor{blue}{69.30})&  90.28(\textcolor{blue}{90.28}) &   66.75(\textcolor{blue}{66.75}) \\    
\bottomrule
\end{tabular}}
\end{table*}

\noindent\textbf{Implementation details.}
We use ResNet-50~\cite{He_CVPR16} as the image encoder and pretrained CLIP~\cite{Radford_ICML21} as the text encoder. Static SOTA methods are trained independently on FashionAI, DeepFashion, and DARN using 100k triplets for 50 epochs, and the best checkpoint is used for inference, strictly following their original official implementations.

In MCL-FIR, all 22 attributes are learned sequentially (FashionAI → DeepFashion → DARN). Each attribute corresponds to an attention module $Attn_i$, trained with 20k \textit{doublets} for 50 epochs before moving to the next. At inference, the shared encoder extracts image features, and the query attribute directly selects the corresponding attention module, eliminating the task-routing required in classical CIL. For the two CIL baselines, 20k \textit{triplets} are sampled per attribute, and all other settings match MCL-FIR for fairness. 

All experiments are conducted on a V100 GPU using PyTorch 2.2.0 with a batch size of 32. Static SOTA methods use an embedding dimension of 1024 and a learning rate of $1\times10^{-4}$ with StepLR. The two CIL baselines and MCL-FIR use $\tau=0.3$ and an embedding dimension of 128. Mean average precision (mAP) is used as the evaluation metric.

\subsection{Main results.}

\noindent\textbf{Main comparisons.} 
Table~\ref{tab:Main_results} compares MCL-FIR with two representative CIL architectures and static baselines, while Table~\ref{tab:time} reports training time under identical settings. Static models must be trained separately on each dataset, whereas the CIL baselines and MCL-FIR follow a sequential protocol that better reflects real-world continual updates.
Across both tables, beyond its scalability, MCL-FIR achieves a strong accuracy–efficiency trade-off. Replacing triplets with InfoNCE doublets removes the need for negative sampling and reduces sampling cost. The multi-head design adds only 0.246M parameters per task, negligible compared to the shared 20M-parameter ResNet-50 backbone. In addition, EMA distillation stabilizes optimization and speeds convergence.

\noindent\textbf{Visualization.}
We randomly sampled 5,000 images and visualized their learned embeddings using t-SNE~\cite{Van_MLR_08} (Fig.~\ref{fig:TSNE}), which shows that MCL-FIR forms well-separated clusters across subclasses. We further visualized attention maps (Fig.~\ref{fig:Attn}), revealing the regions the model focuses on. MCL-FIR consistently highlights discriminative areas for each attribute. For example, for length-related attributes, it attends to both the start and end points of pants, enabling precise length estimation.

\noindent\textbf{Ablation studies.}
We conducted ablation studies to assess each module’s contribution. As shown in Table~\ref{tab:ablation}, replacing the triplet loss with InfoNCE reduces inputs from triplets to doublets and improves performance. Adding the distillation loss $L_{\rm kd}$ further boosts results across all datasets.

\noindent\textbf{Continual learning on the Zappos50K dataset.}
To evaluate MCL-FIR under a more challenging semantic shift, we extend the sequence FashionAI → DeepFashion → DARN by incorporating the footwear dataset Zappos50K~\cite{Yu_CVPR_14}. Unlike apparel, shoes have entirely different attribute concepts, requiring new visual cues and attention adaptations. We follow the standard split in~\cite{Veit_CVPR_17} with 70\%/10\%/20\% for training/validation/testing.

Specifically, after class incremental learning on the three apparel datasets, we further train on Zappos50K and measure both the new-task performance and the forgetting on previous tasks. As shown in Table~\ref{tab:Zappos50K_results}, MCL-FIR maintains consistently high MAP with almost no catastrophic forgetting, and in some cases even improves earlier tasks due to shared fashion-related cues. On Zappos50K, which introduces entirely new semantics, MCL-FIR adapts effectively while preserving prior knowledge, demonstrating an excellent balance between stability and plasticity.

\noindent\textbf{Effect of varying task orders.}
Task order is known to influence CIL performance. We treat each attribute as an individual task and vary only the dataset-level order while keeping intra-dataset attribute order fixed. As shown in Table~\ref{tab:task-order}, performance differences across orders are small, reflecting the stability of our shared encoder and lightweight attribute-specific attention heads. The slight advantage of the FashionAI→DeepFashion→DARN order on FashionAI arises because FashionAI’s fine-grained attributes learned early provide richer representations that benefit later tasks. This indicates dataset-dependent complementarity rather than a limitation of our design, further supporting the robustness of MCL-FIR in practical continual learning settings.

\noindent\textbf{Effect of varying $\lambda$ in Eq.~\ref{eq:final-loss}.}
The variable $\lambda$ is a critical hyperparameter in Eq.~\ref{eq:final-loss}. We conducted experiments to evaluate its influence on retrieval performance by varying its values, and the results are presented in Table~\ref{tab:lamda}. As expected, 
$\lambda$ does not significantly affect the final results. Slightly better performance is achieved when $\lambda$ is set to 0.0001.

\noindent\textbf{Effect of training batch size.}
We examined the impact of batch size $B$ on the instance contrastive loss 
$L_{\rm ins}$. Due to GPU memory limits, batch sizes above 64 were not tested. As shown in Table~\ref{tab:train-batch}, larger batch sizes generally improve retrieval performance by providing more negative samples and enhancing discrimination between pairs.
However, excessively large batches may hinder optimization or weaken gradients.

\noindent\textbf{Effect of the size of training samples.}
We evaluated model performance with different numbers of training doublets, ranging from 2,000 to 100,000, matching the scale used in SOTA methods. As shown in Table~\ref{tab:train-triplets}, even with 2,000 doublets, our model surpasses CSN~\cite{Veit_CVPR_17}. Performance improves with more samples, but excessive data with few epochs can cause underfitting and a slight drop in accuracy.

\noindent\textbf{Training epochs.}
We also examined the impact of training epochs. As shown in Table~\ref{tab:train-epochs}, increasing epochs yields only marginal improvements, indicating that MCL-FIR learns efficiently even with limited training, an advantage for resource-constrained or real-time scenarios.

\noindent\textbf{EMA without distillation.}
We further conducted experiments to investigate whether using only EMA, without distillation, can achieve competitive performance. As shown in Table~\ref{tab:ema-alone}, our method significantly outperforms the EMA-only baseline.

\section{Limitations and Future Works}
While MCL-FIR provides a strong efficiency–accuracy trade-off and supports scalable adaptation to new attributes with substantially reduced training cost, static training can still achieve slightly higher peak accuracy in certain cases. Future work may explore enriching visual representations using additional supervision, as well as incorporating advanced continual-learning strategies such as parameter isolation or dynamic model expansion.

\section{Conclusions}

This paper proposes MCL-FIR, a multi-head continual learning framework for fine-grained FIR.
Through a scalable design that integrates a multi-head architecture, triplet-free contrastive learning, and EMA-guided adaptation, MCL-FIR achieves an excellent balance between efficiency and accuracy.
Thanks to its modular construction, new attributes can be incorporated seamlessly without retraining previously learned components.
Extensive experiments across four datasets, covering both apparel and footwear, demonstrate that MCL-FIR delivers substantial improvements over two representative CIL baselines and achieves performance comparable to static state-of-the-art methods, while requiring only about 30\% of their training cost.
These results indicate that MCL-FIR is both practical and effective for continual fine-grained FIR.

\section*{Acknowledgments}
This research was partially supported by the Japan Society for the Promotion of Science (JSPS) KAKENHI Grant Number 24K20787.

{\small
\bibliographystyle{IEEEtran}
\bibliography{ref}
}

\begin{IEEEbiography}[{\includegraphics[width=1in,height=1.25in,clip,keepaspectratio]{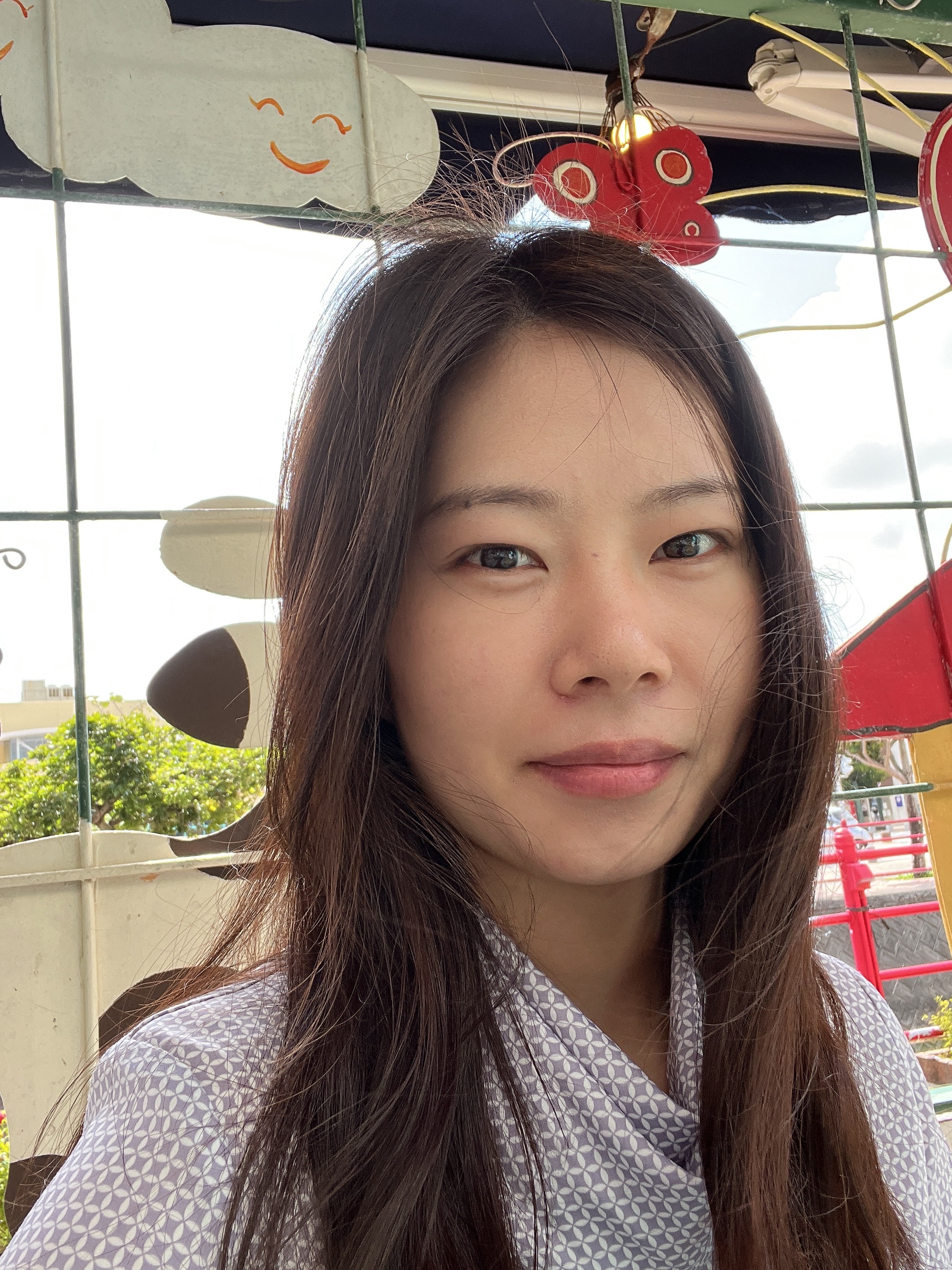}}]{Ling Xiao} (Senior Member, IEEE) received the Ph.D. degree from Huazhong University of Science and Technology in 2020. She is currently an Associate Professor at the Graduate School of Information Science, Hokkaido University, Japan. From October 2023 to March 2025, she served as a Project Assistant Professor at the University of Tokyo. Her current research interests include multi-modal processing, agent AI, continual learning, etc.
\end{IEEEbiography}

\vspace{11pt}

\begin{IEEEbiography}[{\includegraphics[width=1in,height=1.25in,clip,keepaspectratio]{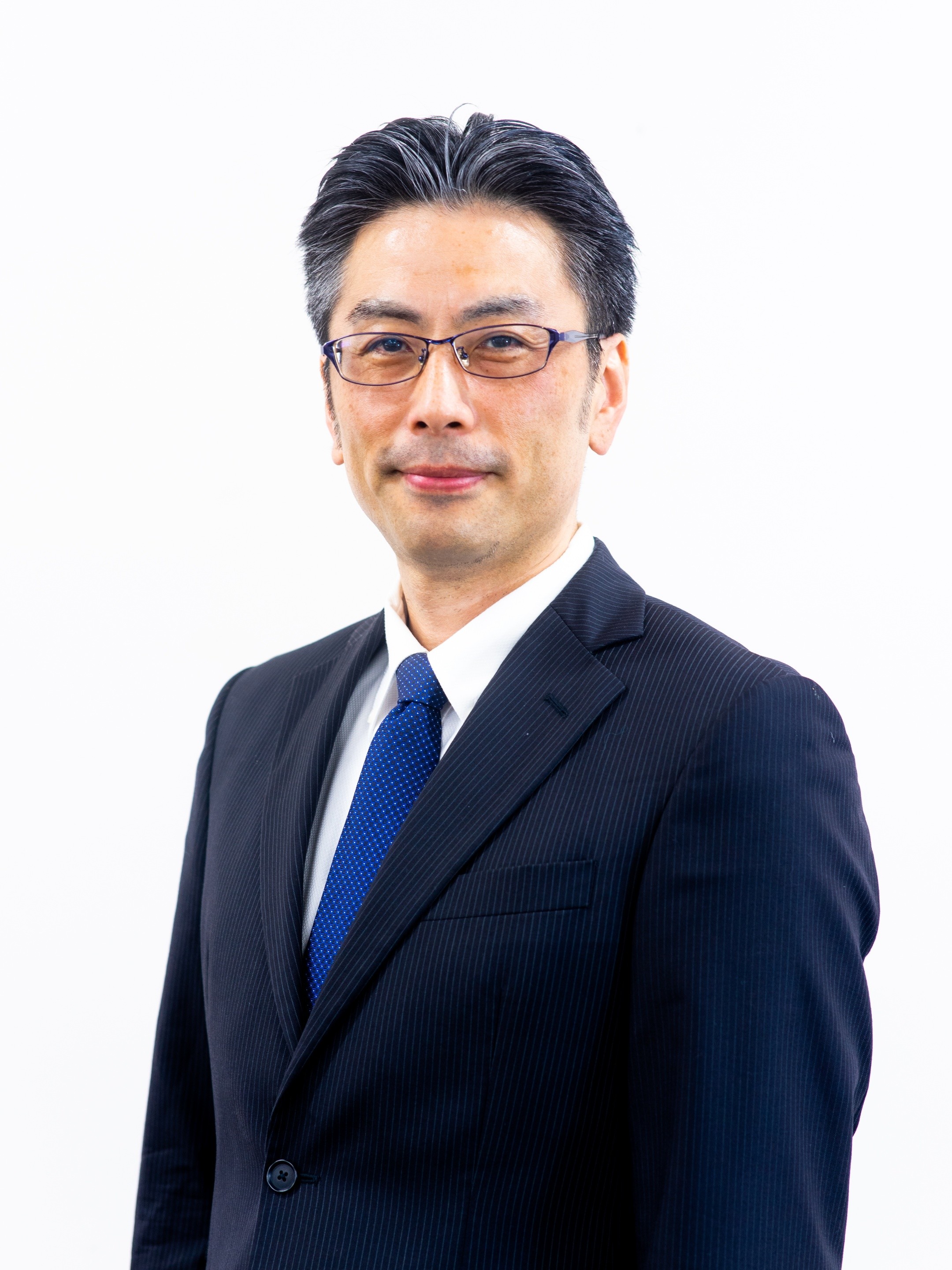}}]{Toshihiko Yamasaki}
(Senior Member, IEEE) received the Ph.D. degree from The University
of Tokyo. He is currently a
Professor with the Department of Information and
Communication Engineering, Graduate School of
Information Science and Technology, The University of Tokyo, Japan. His
current research interests include attractiveness
computing based on multimedia big data analysis, computer vision, pattern
recognition, and machine learning.
\end{IEEEbiography}

\end{document}